\documentclass[a4paper, 10pt, conference]{ieeeconf}      

\usepackage[utf8]{inputenc}
\usepackage{url,hyperref,lineno,microtype,subcaption}
\usepackage{amsmath,amssymb,amsfonts}
\usepackage{algorithmic}
\usepackage{graphicx}
\usepackage{textcomp}
\usepackage{soul}
\usepackage{xcolor}
\usepackage{glossaries}
\usepackage[textwidth=7.25in,textheight=10in]{geometry}
\setacronymstyle{long-short}

\usepackage[backend=biber,bibencoding=utf8,style=ieee,natbib=true]{biblatex}
\IEEEoverridecommandlockouts                              
\title{\LARGE \bf
WaLiN-GUI: a graphical and auditory tool for neuron-based encoding
}

\author{Simon F. M{\"u}ller-Cleve, \textit{EDPR, Istituto Italiano di Tecnologia, Genoa, Italy},\\
Fernando M. Quintana, \textit{School of Engineering, University of Cádiz, Cádiz, Spain},\\
Vittorio Fra, \textit{Politecnico di Torino, Turin, Italy}, \\
Pedro L. Galindo \textit{School of Engineering, University of Cádiz, Cádiz, Spain},\\
Fernando Perez-Peña \textit{School of Engineering, University of Cádiz, Cádiz, Spain},\\
Gianvito Urgese, \textit{Politecnico di Torino, Turin, Italy}, \\
Chiara Bartolozzi, \textit{EDPR, Istituto Italiano di Tecnologia, Genoa, Italy}
\thanks{\small This work is supported by FPU grant (FPU18/04321) from the Spanish Ministry of Universities, the Spanish national research project \textit{NEMOVISION: Sistemas Neuromórficos para Visión Artificial} (PID2019-109465RB-I00), by European Union's Horizon 2020 MSCA Programme under Grant Agreement No. 813713 \textit{NeuTouch}, by the Ebrains-Italy project (CUP B51E22000150006), the 3A-ITALY project (CUP E13C22001900001) and the Fluently project (Grant Agreement No. 101058680).}
}

\begin{document}

\newacronym{bptt}{BPTT}{Backpropagation Through Time}
\newacronym{gui}{GUI}{Graphical User Interface}
\newacronym{snn}{SNN}{Spiking Neural Network}
\newacronym{mn}{MN}{Mihalaş-Niebur}
\newacronym{lif}{LIF}{Leaky Integrate-and-Fire}
\newacronym{ui}{UI}{User Interface}
\newacronym{walin-gui}{WaLiN-GUI}{Watch and Listen into Neurons - Graphical User Interface}

\maketitle
\thispagestyle{empty}
\pagestyle{empty}

\begin{abstract}
Neuromorphic computing relies on spike-based, energy-efficient communication, inherently implying the need for conversion between real-valued (sensory) data and binary, sparse spiking representation. This is usually accomplished using the real valued data as current input to a spiking neuron model, and tuning the neuron's parameters to match a desired, often biologically inspired behaviour. We developed a tool, the WaLiN-GUI, that supports the investigation of neuron models and parameter combinations to identify suitable configurations for neuron-based encoding of sample-based data into spike trains. Due to the generalized LIF model implemented by default, next to the LIF and Izhikevich neuron models, many spiking behaviors can be investigated out of the box, thus offering the possibility of tuning biologically plausible responses to the input data. The GUI is provided open source and with documentation, being easy to extend with further neuron models and personalize with data analysis functions.

\end{abstract}

\section{INTRODUCTION}

The realm of spike-based -- or event-based -- encoding and computation has garnered substantial attention in recent times, evident from the proliferation of publications and advancements in hardware development for sensing~\cite{lichtsteiner2008128,Brandli14,son20174,posch2010qvga} and processing~\cite{davies2018loihi,moradi2017scalable,mayr2019spinnaker,orchard2021efficient}. This surge in interest has mainly focused on event-driven vision, with prominent industry players such as Sony~\cite{Pele} and Intel~\cite{Loihi} venturing into this domain.

Although the terms ``event-based'' and ``spike-based'' are often used interchangeably, we contend that ``event-based'' is better suited to describe artificial systems due to its alignment with the fundamental nature of communication. Unlike clock-driven mechanisms, event-based communication operates asynchronously through discrete binary events that are sparse in time. On the other hand, the term ``spike-based'' finds its place within the realm of neuroscience and biology.

A noteworthy advantage, alongside the attainment of low-latency performance, is the possibility of obtaining reduced energy consumption and memory footprint. Despite the expanding foundation of both software and hardware resources, the event-based sensing capabilities of the research community remain mostly focused in the domain of vision as a consequence of the availability of native event-driven sensors. This constraint arises from multiple factors. One of these is the demanding and time-intensive nature of hardware development, which requires a deep understanding of both design and fabrication processes on the one hand and lossless spike-encoding on the other hand. While in the vision domain it is generally accepted that the event-driven encoding is implemented as a change detection, in other sensory modalities there is a wealth of research to implement conversion with more detailed neuron models, that show closer behaviour to biology, and this changes with the sensory modality. The availability of a tool capable of visualising and measuring the effects of parameters and diverse neuron models with varying encoding schemes is important to support the exploration of encoding strategies, and their effects on spike-based computation, and could potentially catalyze the discovery of novel applications. Addressing these challenges head-on, we introduce the \gls{walin-gui}.

While various \glspl{gui} for modeling neurons and neural networks, offering the ability to manipulate parameters, already exist, none of these effectively facilitate the utilization of recorded experimental data in a tailored manner for the task at hand. Most existing \glspl{gui} serve educational purposes within biology classrooms, while others focus on investigating interactions between groups of neurons. A notable mention is the potent neuron simulator \textit{\href{neuron}{https://neuron.yale.edu/neuron/}}, which allows the definition of diverse neuron models, dynamic adjustment of their parameters, modification of network topology, and a plethora of other functions. This simulation framework strives to replicate \textit{in vivo} cell growth and connectivity development, resulting in high realism but also high complexity. While it provides the essential functions for encoding analysis, users must manually input their custom data. Conversely, the \textit{\href{https://github.com/iankchristie/NeuronModelGUI}{neuronModelGUI}} and \textit{\href{https://github.com/ferper/nesimRT}{NESIM-RT}}~\cite{Rosa-Gallardo23} serves as tools to scrutinize the interactions among multiple neurons of varying types. It enables straightforward, intuitive placement of neurons and connections with adjustable parameters. However, the feasibility of integrating recorded datasets as input to neurons remains uncertain. Moreover, both are coded in Java and C++ respectively, which may impede the incorporation of Python-based neuron models. This is a problem because many machine learning practitioners in the community use Python and leverage PyTorch for training spiking neural networks and none of the aforementioned alternatives supports neuron simulations based on the PyTorch framework, despite some being implemented in Python.

In contrast, our proposed \gls{walin-gui} is designed on purpose to empower developers interested in neuron-based encoding. Its primary goal is to aid in identifying the most suitable neuron model and its associated parameters for converting sequences of sample-based real values into temporally sparse discrete event-based data. Achieving this objective is facilitated through an interactive interface that provides comprehensive information at a glance. Furthermore, the \gls{gui} incorporates auditory inspection, allowing users to assess the impact of neuron parameter adjustments on population firing patterns. To illustrate its efficacy, we utilize example data from the \href{https://zenodo.org/record/7050094}{tactile Braille reading dataset}, acquired using a robotic touch-sensitive fingertip, as presented in~\cite{tactile_braille_reading}. In addition to the well-known Izhikevich~\cite{Izhikevich03} and \gls{lif} neuron models, we introduce the generalized linear \gls{lif} model described by S. Mihalaş and E. Niebur in~\cite{MN} to showcase diverse neuronal responses and the ramifications of distinct neuron parameters. Our \gls{gui} is developed in Python and employs \href{https://www.riverbankcomputing.com/software/pyqt/}{PyQT6} for visualization, \href{http://pydub.com/}{PyDub} for audio output, and \href{https://pytorch.org/}{PyTorch} for neuron modeling. The complete \gls{walin-gui} package is accessible through our Github repository\footnote{\url{https://github.com/event-driven-robotics/WaLiN-GUI}}.

\section{THE WaLiN-GUI}

The \gls{walin-gui} is a Python-based \gls{gui} developed using PyQT6 for visualization, PyTorch for neuron implementations, and PyDub to create the audiotory response. Comprising six primary components, it offers a comprehensive toolset for designing and analyzing spiking neurons. These components are individually described in the following:

\begin{figure*}[htbp]
\includegraphics[width=1.0\textwidth]{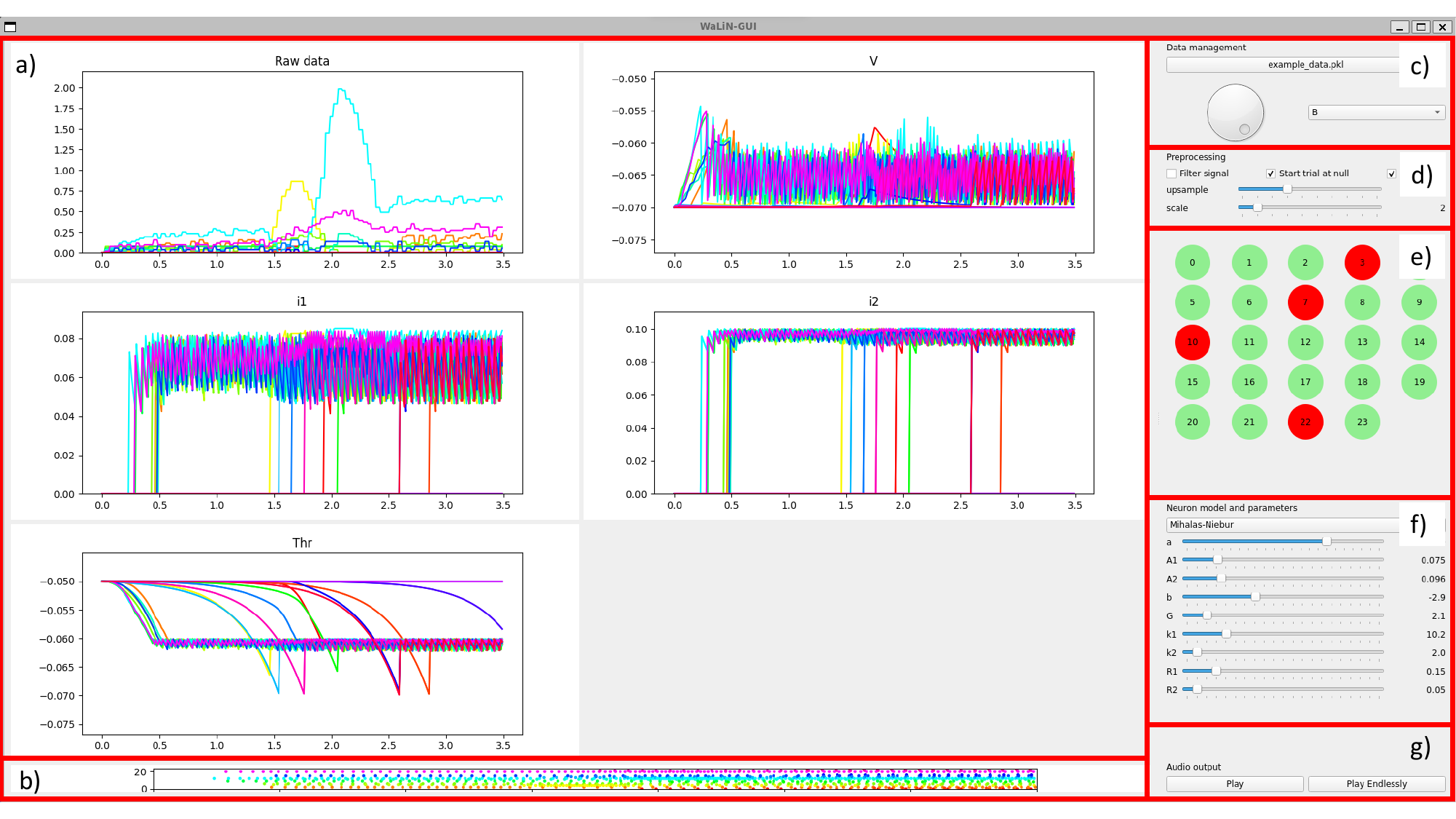}
\caption{The \gls{walin-gui}: on the left-hand side, the plots used to analyze the input data and the neuron dynamics, on the right-hand side, the sample, channels, and parameters settings.}
\label{fig:walin_gui}
\end{figure*}

\subsection{Data Visualization}
This segment presents three critical elements: the raw, sample-based input signal, the internal state variable traces of neurons, and the spike raster plot depicting the network's spiking activity. To increase the numerical stability of the simulation, the data can be sampled by a user-defined factor. To maintain the \gls{gui}'s responsiveness, the visualization remains independent of the up-sampling factor. That is, while updating the neuron model simulation, the visualized data is down-sampled to match the original sampling rate. The display is structured with the sample-based data showcased at the top-left, the neuron membrane potential at the top-right, and the remaining state variables in a $2 \times N$ grid, with $N$ being the remaining included state variables. Furthermore, a compact spike raster plot is positioned at the bottom for rapid spike pattern assessment. For enhanced clarity, the color per channel is synchronized across all plots, simplifying mapping between different visualization elements. The visualization process relies on the Matplotlib library, enabling straightforward adaptation and customization, including the use of various color palettes.

\subsection{Data Format and Selection}
The \gls{walin-gui} currently supports data in the form of a Pandas DataFrame containing a list of individual Python dictionaries, adhering to a specific naming pattern. If the dataset involves multiple classes or repetitions of classes, each trial should contain entries for ``class'' and ``repetition'', if applicable. If both ``class'' and ``repetition'' are applicable, the data selection panel shown in Section c) of Figure~\ref{fig:walin_gui} becomes accessible. The top section displays the loaded file's name, which can be interactively changed using a file browser. The bottom right area houses a combo box for class selection (if multiple classes are indicated by the ``class'' key), while a dial on the left facilitates repetition selection (if multiple repetitions are indicated by the ``repetition'' key).

\subsection{Preprocessing}
Preprocessing options encompass three checkboxes and two sliders. The checkboxes allow for signal filtering using multidimensional image processing from the SciPy library, setting each sensor's starting point to zero, and splitting each sensor channel into two sub-channels, one containing previously positive values and the other containing the absolute values of previously negative readings. This transformation proves advantageous when the data contains negative sensor readings, which could lead to decreases in the neuron's membrane potential, potentially preventing any spiking activity. The two sliders below facilitate up-sampling and up-scaling of the input signal. The sample-based signal is normalized per sensor across all trials. Notably, the simulation time step for neuronal dynamics matches the input data sampling rate and can be adjusted using the up-sample slider.

\subsection{Channel Selection}
The channel selection feature enables users to focus on specific input and encoding channels, simplifying inspection. The sensor layout can be customized to accommodate individual preferences.

\subsection{Neuron Model and Encoding Parameters}
The \gls{walin-gui} extends support to a wide range of neuron models written in Torch, contingent upon adhering to a specific format definition. The currently supported models include the neuron proposed by Mihalas-Niebur in~\cite{MN}, the neuron proposed by Izhikevich~\cite{Izhikevich03}, and the \gls{lif} neuron model. Integration of any model, regardless of its name or number of state variables, is feasible, provided the ``neuronStateVariable'' begins with the neuron's membrane potential voltage ``V'' and ends with the tensor containing the neuron's spikes ``spk''. All intermediary state variables are listed between these two points. Additionally, to facilitate the exchange of manipulable parameters between the neuron simulation and the \gls{gui}, a dictionary named ``parameters\_dict'' must be provided. This dictionary enumerates all parameters intended for manipulation. It is crucial to assign these parameters before running the \gls{gui}, ensuring successful synchronization of neuron parameter updates with slider values. The dictionary includes specifications such as the lowest and highest slider values, step size, and start value.

\subsection{Audio Output}
Positioned at the bottom-right corner, this section governs the audio output of the spike raster plot. The audio output facilitates auditory inspection of the spiking pattern to complement visual examination. Using the PyDub library in Python, the \gls{gui} generates ticks using the sawtooth function, thereby maintaining separation between multiple adjacent ticks and ensuring a distinct auditory experience.

\section{CONCLUSION}
In conclusion, \gls{walin-gui} provides a powerful and versatile interface for researchers and developers working with neuron-based encoding and \glspl{snn}. Its array of features spans data selection, visualization, preprocessing, channel configuration, neuron model integration and tuning, and audio output. This creates a comprehensive toolkit for designing, exploring, and understanding neural encoding by accounting for specific needs depending on the target application. We believe this is a useful addition to the tools available to researchers in the neuromorphic computation and sensing domain and it will grow with the contribution of the community.

\section*{ACKNOWLEDGMENT}
The authors extend their gratitude to the participants and organizers of the 2022 Telluride neuromorphic workshop for fostering invaluable discussions and providing a wellspring of inspiration. A special acknowledgment is reserved for the Neuromorphic Tactile Exploration (NTE) group, whose contributions greatly enriched the workshop experience.


\printbibliography

\end{document}